\documentclass[11pt]{article}
\usepackage[final]{acl}   
\usepackage{times}
\usepackage{latexsym}
\usepackage[T1]{fontenc}
\usepackage[utf8]{inputenc}
\usepackage{microtype}
\usepackage{inconsolata}
\usepackage{graphicx}
\usepackage{booktabs}
\usepackage{tabularx}
\usepackage{amsmath}
\usepackage{xcolor}
\usepackage{colortbl}
\usepackage{listings}
\usepackage{float}
\usepackage{pgfplots}
\pgfplotsset{compat=1.18}

\newcommand{\heat}[1]{%
  \ifdim #1pt>85pt \cellcolor{green!55}%
  \else\ifdim #1pt>70pt \cellcolor{green!28}%
  \else\ifdim #1pt>55pt \cellcolor{yellow!45}%
  \else\ifdim #1pt>40pt \cellcolor{orange!35}%
  \else \cellcolor{red!28}%
  \fi\fi\fi\fi #1}

\lstdefinestyle{promptbox}{
  basicstyle=\ttfamily\scriptsize,
  breaklines=true,
  breakatwhitespace=false,
  breakindent=1em,
  columns=fullflexible,
  keepspaces=true,
  frame=single,
  framerule=0.4pt,
  rulecolor=\color{black!60},
  backgroundcolor=\color{black!4},
  xleftmargin=2pt,
  xrightmargin=2pt,
  aboveskip=6pt,
  belowskip=8pt
}

\title{L3Cube-IndicQuest v2: A Large-Scale Multilingual Benchmark for\\
Evaluating Factual Knowledge of Large Language Models Across Indic Languages}

\author{
  Rinit Jain\textsuperscript{1,2},
  Tirthraj Mahajan\textsuperscript{1,2},
  Advait Joshi\textsuperscript{1,2},
  Raviraj Joshi\textsuperscript{2,3} \\[4pt]
  \textsuperscript{1}Pune Institute of Computer Technology, Pune \\
  \textsuperscript{2}L3Cube Labs, Pune \qquad
  \textsuperscript{3}Indian Institute of Technology Madras \\[4pt]
  \texttt{\{rinitjain9, tirthraj2004, advaitkjoshi, ravirajoshi\}@gmail.com}
}

\begin{document}
\maketitle

\begin{abstract}
We present L3Cube-IndicQuest v2, a large-scale gold-standard multilingual
question-answering benchmark for evaluating the India-specific factual
knowledge of Large Language Models (LLMs). The benchmark comprises 3,471
curriculum-grounded English question--answer pairs spanning nine domains,
curated from educational curricula, competitive examination materials, and
domain-specific reference books. We introduce a practical hybrid construction
strategy that combines context-grounded LLM-based question generation and
validation with semantic deduplication and human verification, enabling
scalable creation of benchmark data while preserving annotation quality. The
benchmark is translated into 19 Indic languages, yielding a publicly released
multilingual dataset of 69,420 question--answer pairs across 20
languages. We evaluate six LLMs under three protocols: LLM-as-a-judge and two deterministic lexical criteria,
exact-substring and word-overlap matching. All three produce almost the same
model ranking, showing that the results do not depend on the choice of judge.
The frontier commercial model leads by a wide margin, and among open-weight
models Gemma4 31B outperforms the Indic-specialised Sarvam 30B in every
evaluated Indic language. 
\end{abstract}

\section{Introduction}

Large Language Models have advanced rapidly, yet their representation of
India-specific knowledge remains weak. Models that do well on general English
benchmarks often answer incorrectly when asked about Indian history, law,
regional culture, or curriculum-level facts~\cite{shafayat2024multifact}. Indic
languages are under-represented in the pre-training corpora of most
multilingual models~\cite{kakwani2020indicnlpsuite, joshi2020state}, and
widely used knowledge benchmarks remain
English-centric~\cite{hendrycks2021measuring}.

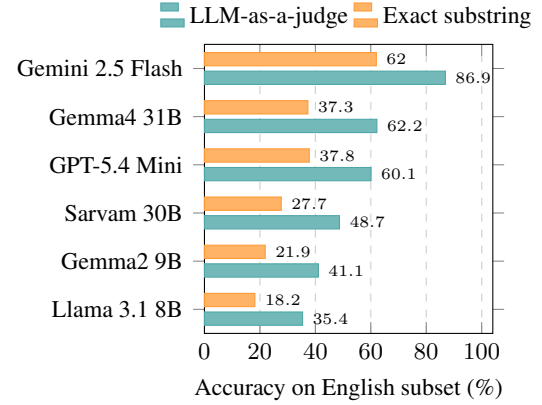
\begin{figure}[!t]
  \centering
  \begin{tikzpicture}
  \begin{axis}[
    xbar, bar width=5pt,
    width=0.70\columnwidth, height=5.4cm,
    xmin=0, xmax=104,
    xlabel={Accuracy on English subset (\%)},
    xlabel style={font=\footnotesize},
    symbolic y coords={Llama 3.1 8B,Gemma2 9B,Sarvam 30B,
                       GPT-5.4 Mini,Gemma4 31B,Gemini 2.5 Flash},
    ytick=data,
    y tick label style={font=\footnotesize},
    x tick label style={font=\footnotesize},
    nodes near coords, nodes near coords style={font=\tiny},
    legend style={font=\footnotesize, at={(0.5,1.02)}, anchor=south,
      legend columns=2, draw=none},
    xmajorgrids=true, grid style={dashed,gray!40},
  ]
  \addplot[fill=teal!55, draw=teal!70] coordinates {
    (86.9,Gemini 2.5 Flash) (62.2,Gemma4 31B) (60.1,GPT-5.4 Mini)
    (48.7,Sarvam 30B) (41.1,Gemma2 9B) (35.4,Llama 3.1 8B)};
  \addplot[fill=orange!60, draw=orange!80] coordinates {
    (62.0,Gemini 2.5 Flash) (37.3,Gemma4 31B) (37.8,GPT-5.4 Mini)
    (27.7,Sarvam 30B) (21.9,Gemma2 9B) (18.2,Llama 3.1 8B)};
  \legend{LLM-as-a-judge, Exact substring}
  \end{axis}
  \end{tikzpicture}
  \caption{Accuracy on the English subset under the LLM judge (Gemma 3 12B)
    and under exact-substring matching. Gemma4 31B and GPT-5.4 Mini swap
    places between the two protocols.}
  \label{fig:teaser}
\end{figure}

Short-form factual question answering is the standard way of measuring such
knowledge. SimpleQA~\cite{wei2024simpleqa} set the format: short questions with
exactly one correct answer, which keeps grading tractable. SimpleQA
Verified~\cite{haas2025simpleqaverified} later showed that such benchmarks
still need deduplication and label checking. Both are English-only, and neither
covers regional knowledge, so neither can measure how well a model knows facts
specific to a country or culture. Benchmarks for Indic languages mostly test
reading comprehension or translation rather than open-domain
recall~\cite{doddapaneni2022indicqa, singh2024indicqabenchmark,
endait2025indicsquad}. L3Cube-IndicQuest v1~\cite{rohera2024indicquest} was the
first to target India-specific factual knowledge across 20 languages, with 200
manually curated pairs per language over five domains, but its scale and
reliance on Wikipedia limited what it could measure.

Scaling that work by hand was not practical. IndicQuest v2 therefore splits the
task: LLM agents do the volume work and human annotators do the judgement work.
One agent generates candidate pairs from chunked curricular text, a second
scores them and cross-checks the answers, a deduplication engine collapses
near-identical questions, and three annotators review everything that survives.
Figure~\ref{fig:teaser} summarises the benchmark and the headline results. Our
contributions are:

\begin{itemize}
  \setlength{\itemsep}{2pt}
  \item A hybrid agent--human pipeline for building curriculum-grounded QA
        data at scale. We combine agentic generation and validation, semantic deduplication, and human verification in a scalable framework for producing high-quality benchmark data.
  \item \textbf{IndicQuest-v2}\footnote{\href{https://huggingface.co/datasets/l3cube-pune/IndicQuest-v2}{l3cube-pune/IndicQuest-v2}}, a publicly released benchmark for
India-specific factual knowledge and hallucination evaluation spanning 20 languages, comprising
3,471 English question--answer pairs across nine domains and 69,420
multilingual pairs in total.
  \item Evidence on the limits of automated validation: human review removed
        20--25\% of pairs that had already passed both automated gates.
  \item Evaluation of six LLMs spanning commercial, general-purpose open-weight, and Indic-specialized models under three protocols: an LLM judge (Gemma 3
        12B), exact-substring matching, and word-overlap matching. The first
        accepts paraphrases; the other two are deterministic and need no judge
        model. The ranking holds across all three.
\end{itemize}

\section{Related Work}

\paragraph{IndicQuest v1.}
L3Cube-IndicQuest v1~\cite{rohera2024indicquest} is the direct predecessor of
this work: 4,000 pairs across English and 19 Indic languages, covering
Literature, History, Geography, Politics, and Economics. With
Llama-3.1-405B-Instruct as judge, English outperformed every Indic language,
and Manipuri, Odia, and Urdu were weakest. Table~\ref{tab:v1v2} summarises how
v2 extends it.

\begin{figure*}[t]
\centering
\scalebox{1}[0.85]{%
  \includegraphics[width=0.88\textwidth]{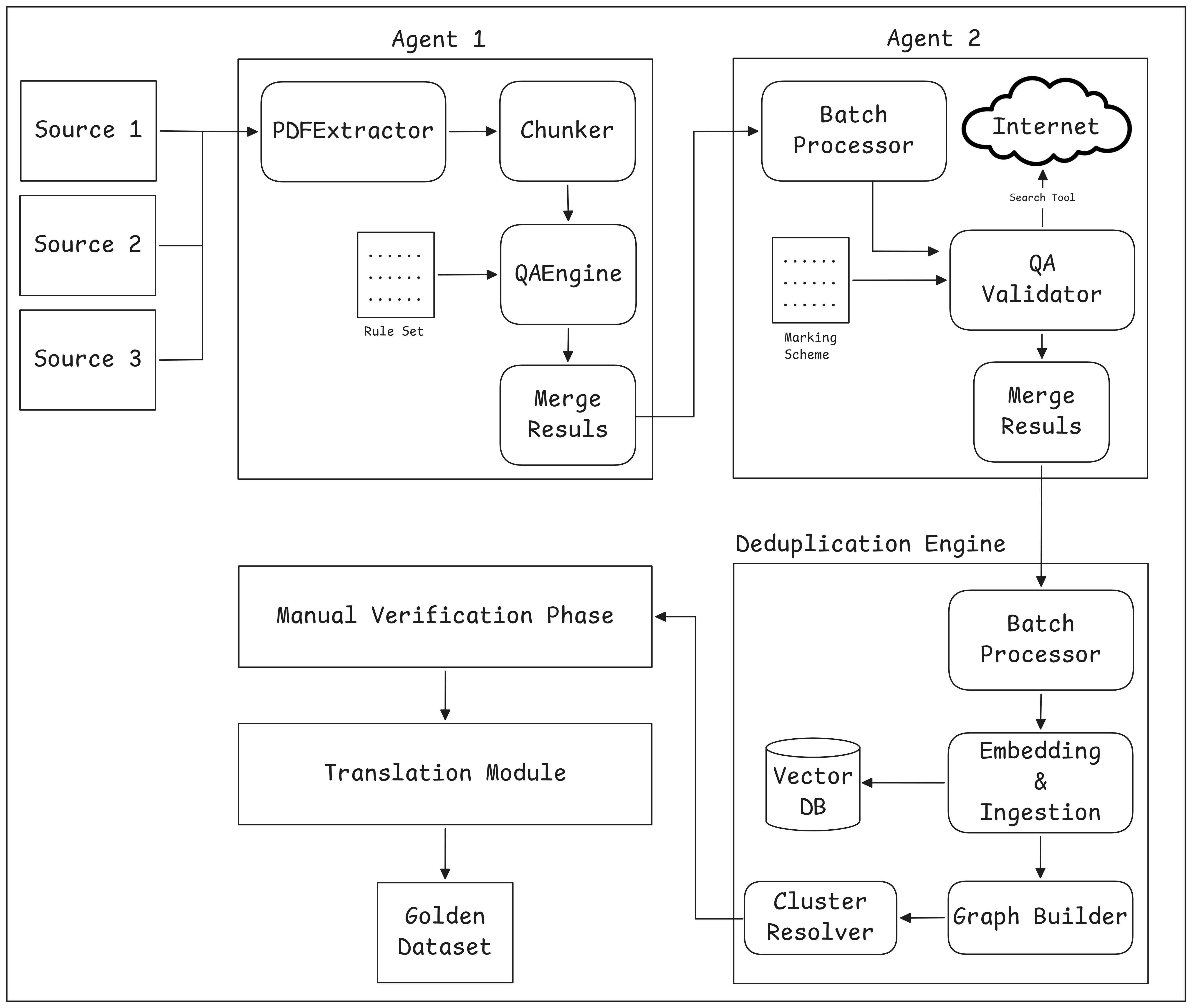}}
\caption{IndicQuest v2 dataset construction pipeline: PDF extraction and
chunking, Agent~1 generation, Agent~2 validation, semantic deduplication,
manual expert review, and translation into 19 Indic languages.}
\label{fig:pipeline}
\end{figure*}

\paragraph{Multilingual QA Benchmarks.}
TyDi QA~\cite{clark2020tydi} covers eleven typologically diverse languages.
XQuAD~\cite{artetxe2020xquad} and MLQA~\cite{lewis2020mlqa} provide
cross-lingual reading comprehension sets derived from
SQuAD~\cite{rajpurkar2016squad}, and MEGA~\cite{ahuja2023mega} spans 16
datasets and 70 languages. All supply a context passage or transfer an
existing task, whereas IndicQuest targets open-domain recall with no passage.

\begin{table}[t]
  \centering
  \small
  \begin{tabularx}{\columnwidth}{lXX}
    \toprule
    \textbf{Attribute} & \textbf{v1} & \textbf{v2} \\
    \midrule
    English pairs   & 200         & 3,471 \\
    Total pairs     & 4,000       & 69,420 \\
    Domains         & 5           & 9 \\
    Source material & Wikipedia, web & Textbooks, UPSC/MPSC, refs \\
    Construction    & Manual      & Hybrid agent--human \\
    Deduplication   & No          & Semantic engine \\
    Verification    & Manual QA   & Three-annotator review \\
    Judge model     & Llama-3.1-405B & Gemma 3 12B \\
    Eval.\ protocols & Judge, ROUGE, F1 & Judge, substring, overlap \\
    Models eval.    & 5           & 6 \\
    Indic model     & No          & Sarvam 30B \\
    \bottomrule
  \end{tabularx}
  \caption{IndicQuest v1 versus v2: key attributes.}
  \label{tab:v1v2}
\end{table}

\paragraph{Short-Form Factuality Benchmarks.}
SimpleQA Verified~\cite{haas2025simpleqaverified} refines
SimpleQA~\cite{wei2024simpleqa} through deduplication and topic balancing,
finding that label noise and redundancy shift measured accuracy. IndicQuest v2
uses the same short-answer design for India-specific knowledge and releases it
in 20 languages.

\paragraph{Culture- and Region-Specific Evaluation.}
BLEnD~\cite{myung2024blend} and INCLUDE~\cite{romanou2024include} test everyday
cultural and regional knowledge. For India, MILU~\cite{verma2024milu} covers
exam-style understanding, IndicGenBench~\cite{singh2024indicgenbench} covers
generation, and PARIKSHA~\cite{watts2024pariksha} studies human--LLM evaluator
agreement. IndicQuest v2 instead grounds open-domain factual QA in Indian
curricular texts.

\paragraph{Indic NLP Datasets and Models.}
IndicQA~\cite{doddapaneni2022indicqa} covers eleven Indic languages using
Wikipedia context paragraphs, \citet{singh2024indicqabenchmark} propose a
closed-form QA benchmark, and IndicSQuAD~\cite{endait2025indicsquad} builds an
extractive set for nine Indic languages from SQuAD. All three evaluate
comprehension over a supplied passage, whereas IndicQuest v2 measures
closed-book recall. IndicNLPSuite~\cite{kakwani2020indicnlpsuite} provides
foundational monolingual corpora, and Airavata~\cite{gala2024airavata} showed
gains from Hindi instruction-tuning, which Sarvam's 30B model extends to
broader Indic pre-training.

\paragraph{LLM-as-a-Judge.}
\citet{zheng2023judging} established the paradigm with MT-Bench and Chatbot
Arena, AlpacaEval~\cite{dubois2024alpacaeval} added length-controlled scoring
to reduce verbosity bias, and \citet{gu2024survey} survey its failure modes.
Since judge quirks can carry into reported scores, we report two deterministic
criteria alongside the judge (Section~\ref{sec:agreement}).

\section{Dataset Construction}

\begin{table*}[t]
  \centering
  \footnotesize
  \setlength{\tabcolsep}{3pt}
  \begin{tabularx}{\textwidth}{lXl}
    \toprule
    \textbf{Domain} & \textbf{Question} & \textbf{Answer} \\
    \midrule
    Art & Which specific architectural element of Sanchi Stupa-1 features four
      beautifully decorated structures depicting Jatakas? & Toranas \\
    Commercial Studies & Which apex body was set up in 1982 to coordinate all
      institutions involved in the rural financing system? & NABARD \\
    Culture & Which specific folk song is sung in Chhattisgarh to propitiate
      Lord Shiva and Goddess Durga? & Goura geet \\
    Geography & Which river's water is primarily harnessed by the Ukai project
      in India? & Tapi river \\
    History & What specialised weapon did Shivaji Maharaj first thrust into
      Afzal Khan's stomach during their embrace? & Waghnakh \\
    Law & When did the insertion of section 29A, defining ``electronic
      record'', by Act 21 of 2000 come into effect? & 17-10-2000 \\
    Political Science & What was the original number of seats allocated to the
      Princely States in the Indian Constituent Assembly under the Cabinet
      Mission Plan? & 93 seats \\
    Science & Which umbrella-shaped astronomical device does the Arya Siddhanta
      describe? & Chatra yantra \\
    Sports & Which is the oldest football tournament in India, established by
      Sir Mortimer Durand and held from 1888 in Simla? & Durand Cup \\
    \bottomrule
  \end{tabularx}
  \caption{Representative pairs from the IndicQuest v2 English subset, one per
    domain. Every gold answer is one to five words and appears verbatim in the
    source text.}
  \label{tab:examples}
\end{table*}

\subsection{Source Material}

IndicQuest v2 draws on formal Indian educational and reference texts, a
deliberate shift from the web sources used in v1:

\begin{itemize}
  \setlength{\itemsep}{2pt}
  \item \textbf{School curricula:} NCERT (Classes 6--12), SSC Maharashtra Board
        (Classes 4--10), CBSE, ICSE, and HSC textbooks in History, Geography,
        Science, and Social Studies.
  \item \textbf{Competitive examination material:} UPSC and MPSC preparation
        books covering Indian polity, history, economics, and general
        knowledge.
  \item \textbf{Domain references:} books on Indian law and the Constitution,
        and specialised texts on the history of Indian science and sport.
\end{itemize}

Indian sports history and Indian contributions to science are largely absent
from school curricula, making them the most novel domains in v2. Construction

\subsection{Question Design Principles}
\label{sec:design}

Three constraints govern every question, and both agents enforce them.

\paragraph{Short answers.}
A short answer is one to five words long and appears verbatim in the source
text: a name, a date, a place, or a term, never a sentence or an explanation.
This makes evaluation simpler. A short phrase either matches the gold reference
or it does not, so there is little of the ambiguity that comes with comparing
long free-form answers, and the responses can also be scored by plain string
comparison with no judge model (Section~\ref{sec:lexical}).

\paragraph{Non-ambiguity.}
Each question is written so that exactly one answer is correct. The two
pipeline agents described in Section~\ref{sec:pipeline} enforce this from both
ends: the generation agent avoids vague or directional phrasings and uses
complete proper names, and the validation agent rejects any pair admitting more
than one defensible answer. Ambiguous pairs are discarded rather than repaired.

\paragraph{Self-containment and India specificity.}
Questions must be answerable without the source passage and must not refer to
it, since no context is supplied at evaluation time. Every question must
concern Indian history, geography, polity, law, culture, science, or sport.
Table~\ref{tab:examples} gives one example per domain.

\subsection{Construction Pipeline}
\label{sec:pipeline}

The principles above are enforced by a six-stage pipeline
(Figure~\ref{fig:pipeline}); prompts are in Appendix~\ref{app:prompts}.

\begin{itemize}
  \setlength{\itemsep}{0pt}
  \setlength{\parskip}{0pt}
  \setlength{\topsep}{2pt}
  \item \textbf{Extraction:} source books converted to text, split into
        passage-sized chunks.
  \item \textbf{Agent~1:} generates candidate pairs from each chunk.
  \item \textbf{Agent~2:} scores each pair and cross-checks the answer.
  \item \textbf{Deduplication:} collapses near-identical questions from
        overlapping books.
  \item \textbf{Verification:} three annotators review every surviving pair.
  \item \textbf{Translation:} the verified English set goes into 19 Indic
        languages.
\end{itemize}

\subsection{Agent 1: Question--Answer Generation}

Agent~1 runs on Gemini 2.5 Flash, taking chunked text from the source PDFs and
generating factual, self-contained pairs. Its prompt
(Appendix~\ref{app:agent1}) enforces the constraints of
Section~\ref{sec:design}, plus enough difficulty to separate models.

\subsection{Agent 2: Quality Validation}

Agent~2, also on Gemini 2.5 Flash (Appendix~\ref{app:agent2}), scores every
candidate on nine dimensions (0--10 each, 90 total): Correctness, Relevance,
Difficulty, Overall Quality, India Specificity, Context Richness, Answer
Quality, Grammar and Structure, and Specialised Knowledge. It also sets a
binary \texttt{correctness\_flag} and a \texttt{google\_search\_confidence}
rating from external sources. Pairs flagged FALSE are dropped; the rest are
ranked by score and passed on.

\subsection{Semantic Deduplication}

Books covering the same topic yield equivalent questions in different words.
The engine clusters ranked pairs by sentence-embedding
similarity~\cite{reimers2019sentence} and keeps the highest-scoring member of
each cluster. Table~\ref{tab:dedup} shows a real case: four questions about
Qutbuddin Aibak, from four books, cluster together and only the top one
survives. Most clusters are singletons and pass through unchanged.

\begin{table*}[t]
  \centering
  \footnotesize
  \setlength{\tabcolsep}{2.5pt}
  \renewcommand{\arraystretch}{1.15}
  \begin{tabularx}{\textwidth}{lXlcl}
    \toprule
    \textbf{Status} & \textbf{Question} & \textbf{Answer} & \textbf{Score} &
    \textbf{Source} \\
    \midrule
    \multicolumn{5}{l}{\textit{Cluster 19 --- Brahmi script (4 variants)}} \\
    \rowcolor{green!16}
    Retained & What type of script was used for the twelve Ashokan inscription
      findspots that appeared in South India? & Brahmi script & 90 & NCERT
      Class 7 History \\
    \rowcolor{red!10}
    Discarded & In which script were the majority of Ashokan inscriptions
      composed across the greater part of the Indian subcontinent? & Brahmi
      script & 88 & SSC History Class 9 \\
    \rowcolor{red!10}
    Discarded & Which ancient Indian script was used for the short inscriptions
      found in Sri Lanka dating to the second and first centuries BC? & Brahmi
      script & 85 & UPSC \\
    \rowcolor{red!10}
    Discarded & Which script did Ashoka primarily use for writing the majority
      of his royal messages and inscriptions? & Brahmi script & 81 & MPSC
      History \\
    \midrule
    \multicolumn{5}{l}{\textit{Cluster 20 --- single representative, no duplicates found}} \\
    \rowcolor{green!16}
    Retained & Which Sultan of Delhi politely refused asylum to Jalaluddin when
      Changez Khan was near the Indus? & Iltutmish & 90 & NCERT Class 7 \\
    \bottomrule
  \end{tabularx}
  \caption{Semantic deduplication in practice. Green rows are retained as
    cluster representatives; red rows are discarded duplicates. Cluster~19
    collapses four equivalent questions drawn from four source books;
    Cluster~20 shows the common singleton case, which passes unchanged to
    manual verification.}
  \label{tab:dedup}
\end{table*}

\subsection{Domain Split and Manual Verification}

The English dataset then contains 3,471 pairs across nine domains: Culture
(534), History (510), Geography (499), Law (494), Political Science (450),
Science (325), Sports (273), Art (233), and Commercial Studies (153). Sizes
reflect the source material available. Three annotators next reviewed every pair independently, flagging factual
errors, ambiguous phrasing, trivially answerable content, and weak India
specificity. Any pair flagged by one reviewer was removed after discussion.
This cut 20--25\% of the set: a share of Agent~1's output passed every
automated check but failed expert review.

\subsection{Translation and Release}

The verified pairs were translated into 19 Indic languages: Assamese, Bengali,
Dogri, Gujarati, Hindi, Kannada, Konkani, Maithili, Malayalam, Marathi, Meitei
(Manipuri), Nepali, Odia, Punjabi, Sanskrit, Sindhi, Tamil, Telugu, and Urdu,
using GPT-5.4 Mini via batched API calls. Google Cloud and Gemini Translation
were tried first but handled Law and Political Science terminology
inconsistently. Questions and answers went in separate batches with no pairing information, so
the model never saw factual associations and translation stayed purely
linguistic. The sets were merged afterwards.

\section{Evaluation Methodology}

\begin{table*}[t]
  \centering
  \footnotesize
  \renewcommand{\arraystretch}{1.15}
  \setlength{\tabcolsep}{4pt}
  \begin{tabular}{lcccccc}
    \toprule
    \textbf{Domain} &
    \textbf{Gemini 2.5 Flash} &
    \textbf{Gemma4 31B} &
    \textbf{GPT-5.4 Mini} &
    \textbf{Sarvam 30B} &
    \textbf{Gemma2 9B} &
    \textbf{Llama 3.1 8B} \\
    \midrule
    Art (233)               & \heat{75.5} & \heat{70.0} & \heat{65.2} & \heat{48.5} & \heat{41.2} & \heat{30.9} \\
    Commercial (153)        & \heat{94.8} & \heat{75.2} & \heat{72.5} & \heat{67.3} & \heat{56.2} & \heat{55.6} \\
    Culture (534)           & \heat{75.7} & \heat{65.2} & \heat{61.0} & \heat{44.4} & \heat{39.7} & \heat{31.3} \\
    Geography (499)         & \heat{87.6} & \heat{59.3} & \heat{60.5} & \heat{45.9} & \heat{42.5} & \heat{36.9} \\
    History (510)           & \heat{95.3} & \heat{65.1} & \heat{60.2} & \heat{48.6} & \heat{36.7} & \heat{31.0} \\
    Law (494)               & \heat{89.7} & \heat{50.8} & \heat{45.1} & \heat{31.2} & \heat{26.9} & \heat{29.4} \\
    Political Science (450) & \heat{92.9} & \heat{72.7} & \heat{73.8} & \heat{67.8} & \heat{56.4} & \heat{47.3} \\
    Science (325)           & \heat{88.6} & \heat{68.3} & \heat{67.4} & \heat{64.6} & \heat{53.2} & \heat{42.2} \\
    Sports (273)            & \heat{81.0} & \heat{38.1} & \heat{41.8} & \heat{34.1} & \heat{27.1} & \heat{24.5} \\
    \midrule
    \textbf{Weighted}       & \heat{86.9} & \heat{62.2} & \heat{60.1} & \heat{48.7} & \heat{41.1} & \heat{35.4} \\
    \bottomrule
  \end{tabular}
  \caption{Domain-wise model accuracy on the English subset (\%) under the LLM
    judge, with domain sizes in parentheses. Shading: green $>$ 85\%, light
    green 70--85\%, yellow 55--70\%, orange 40--55\%, red $<$ 40\%.
    Corresponding lexical-criterion scores are given in
    Table~\ref{tab:lexical_domains}.}
  \label{tab:heatmap}
\end{table*}

\subsection{Models Evaluated}

Six LLMs were selected to span a range of scales, architectures, and training
philosophies:

\begin{itemize}
  \setlength{\itemsep}{2pt}
  \item \textbf{Gemini 2.5 Flash} (Google): frontier reference model. It is
        also the Agent~1 and Agent~2 backbone, which makes its score a data
        point on generator--evaluator overlap.
  \item \textbf{GPT-5.4 Mini} (OpenAI): a general-purpose model testing
        whether broad multilingual pre-training transfers to India-specific
        factual knowledge.
  \item \textbf{Gemma4 31B} (Google): an open-weight model, included to
        test whether open-weight systems at the 30B scale can challenge
        proprietary ones.
  \item \textbf{Sarvam 30B}: an Indic-native model pre-trained on
        Indian-language corpora, included to test whether Indic-specific
        pre-training yields an advantage.
  \item \textbf{Gemma2 9B}~\cite{team2024gemma}: a mid-sized model
        representative of deployments in resource-constrained settings.
  \item \textbf{Llama 3.1 8B}~\cite{grattafiori2024llama}: a community
        baseline establishing the performance floor.
\end{itemize}

\subsection{LLM-as-a-Judge Evaluation}

Each model was prompted with each question without retrieval or context
augmentation and instructed to answer in the language of the question
(Appendix~\ref{app:qa_prompt}). Responses were evaluated by Gemma 3 12B
(Appendix~\ref{app:judge_prompt}), which receives the question, the gold answer
and the model response, and returns a binary verdict with a natural-language
justification. The judge accepts paraphrases and alternate transliterations:
``Chhatrapati Shivaji Maharaj'' and ``Shivaji Raje Bhosle'' are both correct
answers to \emph{What was the name of the king who founded the Maratha
Empire?} Responses in a language other than that of the question are marked
incorrect, penalising language inconsistency. The pipeline is shown in
Figure~\ref{fig:eval}.

\subsection{Deterministic Lexical Evaluation}
\label{sec:lexical}

Because every gold answer is one to five words, the benchmark can also be
scored without a judge. We report two deterministic criteria alongside the
judge verdicts, applied to the same 3,471 responses per model:

\begin{itemize}
  \setlength{\itemsep}{2pt}
  \item \textbf{Exact substring:} the complete gold answer must occur in the
        response as one contiguous, case-insensitive substring.
  \item \textbf{Word overlap:} every word of the gold answer must occur
        somewhere in the response, in any order, case-insensitively and after
        stripping edge punctuation.
\end{itemize}

Neither criterion subsumes the other: substring matching ignores word
boundaries, while word overlap ignores order but demands exact tokens. Both are
stricter than the judge and fully reproducible from the released responses.

\subsection{Metrics}

Domain accuracy is the fraction of correct verdicts among the $|D_d|$ questions
of domain $d$, under whichever protocol is in use. Weighted accuracy aggregates
across domains in proportion to domain size:
\begin{equation}
  \text{Weighted Accuracy} = \frac{\displaystyle\sum_{d} |D_d| \cdot
    \text{Accuracy}_d}{\displaystyle\sum_{d} |D_d|}
\end{equation}

\section{Results and Analysis}
\subsection{English Results}

Table~\ref{tab:heatmap} gives domain-wise judge accuracy for all six models on
the English subset. The overall ordering is:

\begin{itemize}
  \setlength{\itemsep}{1pt}
  \item \textbf{Gemini 2.5 Flash}: 86.9\% (3,018 of 3,471 correct)
  \item \textbf{Gemma4 31B}: 62.2\% (2,158)
  \item \textbf{GPT-5.4 Mini}: 60.1\% (2,086)
  \item \textbf{Sarvam 30B}: 48.7\% (1,692)
  \item \textbf{Gemma2 9B}: 41.1\% (1,427)
  \item \textbf{Llama 3.1 8B}: 35.4\% (1,228)
\end{itemize}

This ordering is stable but not uniform: GPT-5.4 Mini overtakes Gemma4 31B in
Geography, Political Science, and Sports, and Llama 3.1 8B overtakes Gemma2 9B
in Law. Gemini's lead should be read alongside its role as the Agent~1 and
Agent~2 backbone (see Limitations).

\begin{table}[t]
  \centering
  \small
  \setlength{\tabcolsep}{4pt}
  \begin{tabular}{lccc}
    \toprule
    \textbf{Model} &
    \shortstack{\textbf{LLM as}\\\textbf{a judge}} &
    \shortstack{\textbf{Exact substring}\\\textbf{matching}} &
    \shortstack{\textbf{Word}\\\textbf{overlap}} \\
    \midrule
    Gemini 2.5 Flash & 86.9 & 62.0 & 62.2 \\
    Gemma4 31B       & 62.2 & 37.3 & 37.5 \\
    GPT-5.4 Mini     & 60.1 & 37.8 & 37.5 \\
    Sarvam 30B       & 48.7 & 27.7 & 27.9 \\
    Gemma2 9B        & 41.1 & 21.9 & 21.4 \\
    Llama 3.1 8B     & 35.4 & 18.2 & 17.9 \\
    \bottomrule
  \end{tabular}
  \caption{Accuracy (\%) on the English subset under all three evaluation
    protocols.}
  \label{tab:agreement}
\end{table}

\begin{table*}[t]
  \centering
  \footnotesize
  \renewcommand{\arraystretch}{1.05}
  \setlength{\tabcolsep}{6pt}
  \begin{tabular}{lrcccccccccc}
    \toprule
    \textbf{Domain} & \textbf{Total} &
    \textbf{Eng} & \textbf{Guj} & \textbf{Hin} & \textbf{Ass} & \textbf{Tel} &
    \textbf{Odi} & \textbf{San} & \textbf{Mar} & \textbf{Ben} & \textbf{Tam} \\
    \midrule
    \multicolumn{12}{l}{\textit{(a) Gemma4 31B}} \\
    Art              & 233 & 70.0 & 65.2 & 70.4 & 69.3 & 68.5 & 68.3 & 66.3 & 70.4 & 66.0 & 64.1 \\
    Commercial       & 153 & 75.2 & 68.0 & 69.9 & 70.1 & 69.3 & 69.1 & 67.1 & 66.0 & 66.8 & 64.9 \\
    Culture          & 534 & 65.2 & 63.9 & 69.9 & 66.6 & 65.7 & 65.6 & 63.5 & 66.3 & 63.1 & 61.2 \\
    Geography        & 499 & 59.3 & 52.1 & 63.5 & 59.8 & 58.9 & 58.7 & 56.5 & 63.1 & 56.1 & 54.1 \\
    History          & 510 & 65.1 & 59.2 & 68.4 & 63.1 & 62.2 & 62.0 & 59.8 & 58.2 & 59.4 & 57.5 \\
    Law              & 494 & 50.8 & 43.1 & 62.3 & 51.2 & 50.3 & 50.1 & 47.8 & 47.4 & 47.4 & 45.4 \\
    Political Sci.   & 450 & 72.7 & 71.8 & 80.2 & 74.4 & 73.7 & 73.5 & 71.8 & 72.0 & 71.4 & 69.7 \\
    Science          & 325 & 68.3 & 63.1 & 69.5 & 66.5 & 65.7 & 65.5 & 63.4 & 64.0 & 63.1 & 61.1 \\
    Sports           & 273 & 38.1 & 42.9 & 43.6 & 41.6 & 40.7 & 40.5 & 38.3 & 40.7 & 37.9 & 36.0 \\
    \textbf{Weighted} & -- & \textbf{62.2} & \textbf{58.1} & \textbf{66.9} &
      \textbf{62.3} & \textbf{61.4} & \textbf{61.2} & \textbf{59.1} &
      \textbf{60.7} & \textbf{58.8} & \textbf{56.9} \\
    \midrule
    \multicolumn{12}{l}{\textit{(b) Sarvam 30B}} \\
    Art              & 233 & 48.5 & 45.1 & 41.2 & 44.6 & 42.9 & 40.8 & 38.6 & 40.3 & 38.2 & 33.9 \\
    Commercial       & 153 & 67.3 & 56.2 & 58.8 & 54.2 & 53.6 & 53.3 & 51.0 & 47.7 & 49.0 & 47.1 \\
    Culture          & 534 & 44.4 & 40.8 & 36.0 & 38.6 & 36.9 & 36.3 & 34.2 & 34.1 & 35.0 & 33.0 \\
    Geography        & 499 & 45.9 & 35.5 & 34.9 & 31.7 & 34.3 & 32.6 & 30.6 & 29.9 & 29.1 & 27.9 \\
    History          & 510 & 48.6 & 41.0 & 39.8 & 39.0 & 37.5 & 38.0 & 35.9 & 37.6 & 32.9 & 36.1 \\
    Law              & 494 & 31.2 & 33.6 & 28.3 & 35.0 & 25.7 & 29.0 & 27.2 & 26.9 & 32.6 & 26.5 \\
    Political Sci.   & 450 & 67.8 & 52.0 & 50.4 & 48.0 & 52.0 & 49.7 & 47.5 & 45.6 & 47.3 & 42.9 \\
    Science          & 325 & 64.6 & 47.4 & 45.8 & 41.8 & 43.7 & 43.5 & 41.3 & 38.2 & 36.9 & 38.2 \\
    Sports           & 273 & 34.1 & 23.4 & 24.9 & 21.2 & 21.6 & 22.6 & 21.1 & 22.7 & 20.5 & 19.8 \\
    \textbf{Weighted} & -- & \textbf{48.7} & \textbf{40.7} & \textbf{38.6} &
      \textbf{38.4} & \textbf{37.6} & \textbf{37.4} & \textbf{35.3} &
      \textbf{35.0} & \textbf{35.0} & \textbf{33.2} \\
    \bottomrule
  \end{tabular}
  \caption{Judge accuracy (\%) of Gemma4 31B and Sarvam 30B across domains and
    languages. Eng = English, Guj = Gujarati, Hin = Hindi, Ass = Assamese,
    Tel = Telugu, Odi = Odia, San = Sanskrit, Mar = Marathi, Ben = Bengali,
    Tam = Tamil.}
  \label{tab:multiling}
\end{table*}

\subsection{Agreement Between Evaluation Protocols}
\label{sec:agreement}

Table~\ref{tab:agreement} compares the judge against the two deterministic
criteria. Three results follow.

\begin{itemize}
  \setlength{\itemsep}{4pt}

  \item \textbf{The ranking does not depend on the protocol.}
  The three protocols give almost the same ordering, and the two lexical criteria
  differ by at most 0.5 pp on any model. The ranking is a property of the
  models, not of the judge we chose.

  \item \textbf{Gemma4 31B matches GPT-5.4 Mini.}
  The judge puts Gemma4 ahead by 2.10 pp, while exact substring puts GPT-5.4
  Mini ahead by 0.43 pp and word overlap by 0.09 pp. No protocol separates
  them, so an open-weight model here almost matches a commercial one.

  \item \textbf{Lexical scores are a strict floor.}
  Every model scores 17--25 pp lower under string matching, and the shortfall
  is uneven. It is smallest in Law and Sports, whose answers are dates and
  fixed English names such as \emph{Durand Cup}, and largest where answers are
  Indic words such as \emph{Waghnakh} and \emph{Toranas}, which have more than
  one accepted spelling (Ex: \emph{Vagh Nakh}, \emph{Toran}). String matching therefore under-reports the knowledge
  the benchmark measures, so we report the LLM judge as the primary protocol and the lexical
scores as a reproducible floor.

\end{itemize}

\subsection{Multilingual Results}

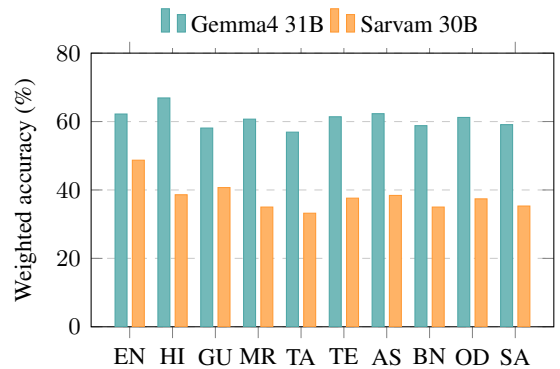
\begin{figure}[!b]
  \centering
  \begin{tikzpicture}
  \begin{axis}[
    ybar, bar width=4.5pt,
    width=\columnwidth, height=5.2cm,
    ymin=0, ymax=80,
    ylabel={Weighted accuracy (\%)},
    ylabel style={font=\footnotesize},
    symbolic x coords={EN,HI,GU,MR,TA,TE,AS,BN,OD,SA},
    xtick=data,
    x tick label style={font=\footnotesize},
    y tick label style={font=\footnotesize},
    legend style={font=\footnotesize, at={(0.5,1.02)}, anchor=south,
      legend columns=2, draw=none},
    ymajorgrids=true, grid style={dashed,gray!40},
  ]
  \addplot[fill=teal!55, draw=teal!70] coordinates {
    (EN,62.2)(HI,66.9)(GU,58.1)(MR,60.7)(TA,56.9)
    (TE,61.4)(AS,62.3)(BN,58.8)(OD,61.2)(SA,59.1)
  };
  \addplot[fill=orange!60, draw=orange!80] coordinates {
    (EN,48.7)(HI,38.6)(GU,40.7)(MR,35.0)(TA,33.2)
    (TE,37.6)(AS,38.4)(BN,35.0)(OD,37.4)(SA,35.3)
  };
  \legend{Gemma4 31B, Sarvam 30B}
  \end{axis}
  \end{tikzpicture}
  \caption{Weighted accuracy of Gemma4 31B and Sarvam 30B across ten
    languages, under the LLM-as-a-judge Gemma3 12B.}
  \label{fig:lang_compare}
\end{figure}

Table~\ref{tab:multiling} reports Gemma4 31B and Sarvam 30B, the two same-scale
models, across all nine domains in ten languages;
Figure~\ref{fig:lang_compare} plots their weighted accuracy side by side.
Gemma4 31B leads in every language, and its spread across languages
(56.9--66.9\%) is far narrower than Sarvam's (33.2--48.7\%).

\subsection{Key Observations}
\label{sec:key_obs}

\begin{enumerate}
  \setlength{\itemsep}{4pt}

\item \textbf{Indic-native pre-training does not guarantee an advantage (Gemma4 31B vs Sarvam 30B).}
The comparison shows a substantial performance difference between a dense model (Gemma4 31B) and a mixture-of-experts model (Sarvam 30B), despite their similar nominal parameter scales:
\begin{itemize}
  \setlength{\itemsep}{0pt}
  \setlength{\parskip}{0pt}
  \setlength{\topsep}{2pt}
  \item Gemma4 31B reaches 62.2\% against Sarvam 30B's 48.7\%: a gap of
        13.5 pp, or 466 questions.
  \item The ordering holds under both lexical criteria (37.3\% versus 27.7\%
        on exact substring).
  \item Gemma4's largest margins are Art (21.5 pp), Culture (20.8 pp), Law
        (19.6 pp), and History (16.5 pp).
  \item Sarvam 30B comes closest in Science (3.7 pp), Sports (4.0 pp), and
        Political Science (4.9 pp), but leads in no domain.
  \item Gemma4 31B leads in every language, and the gap widens to 17.4--28.3 pp
        across Indic languages, peaking in Hindi.
\end{itemize}
This result indicates that Indic-specific training does not by itself guarantee an advantage, although the architectural and training differences between the two models prevent attributing the gap to pre-training strategy alone.

\item \textbf{Parameter count no longer predicts rank cleanly.}
Gemini 2.5 Flash leads every domain and the English ranking by 24.8 pp,
matching v1's finding that frontier models dominate. Below it, larger is not
always better: the open-weight Gemma4 31B almost ties with GPT-5.4 Mini
(Section~\ref{sec:agreement}), and Llama 3.1 8B beats the larger Gemma2 9B in
Law. Architecture and training data matter as much as parameter count.

\item \textbf{English supremacy is no longer universal.}
In v1, English beat every Indic language for every model. In v2 that holds for
Sarvam 30B but not Gemma4 31B, where Hindi (66.9\%) and Assamese (62.3\%) edge
out English (62.2\%). English is no longer a guaranteed ceiling, though most
Indic languages still trail.

\item \textbf{Domain hierarchy: Sports and Law are the weak points.}
Averaged over all six models, Commercial Studies is strongest ($\sim$70\%),
then Political Science ($\sim$68\%) and Science ($\sim$64\%), while Sports
($\sim$41\%) and Law ($\sim$46\%) are weakest. In v1 Geography was weakest.
Both stay lowest in every language for both models
(Table~\ref{tab:multiling}), so procedural and fast-changing knowledge appear
under-represented in pre-training corpora.

\item \textbf{Language-level performance is model-dependent.}
Tamil is weakest for both Sarvam 30B (33.2\%) and Gemma4 31B (56.9\%), but the
second-weakest differs: Marathi and Bengali for Sarvam, Gujarati for Gemma4.
Performance depends on each model's pre-training corpus, not the language
alone.

\item \textbf{Hallucination in curriculum-grounded generation.}
The 20--25\% manual removal rate shows how often a frontier model, given source
text and checked by a second automated pass, still produces pairs that fail
expert review. It sets a lower bound on the oversight such pipelines need.
\end{enumerate}

\section{Future Work}

Two extensions are planned. First, a region-specific structure: v2 translates
one English pool into all 19 languages, whereas an 80--20 split, with 20\% of
questions curated natively per region, would test regional depth too. Example, the
20\% of Marathi dataset would cover Maharashtra-specific sources to gain more cultural context. Second, extending the evaluation to all 19
languages and all six models under all three protocols. Since string matching
penalises spelling variants (Section~\ref{sec:agreement}), a
spelling-normalised criterion would give a fairer judge-free floor for
Indic-script responses.

\section{Conclusion}

We present L3Cube-IndicQuest v2, a benchmark for evaluating LLM knowledge in
the Indian context. Built through a two-agent Gemini 2.5 Flash pipeline,
semantic deduplication, and human verification, it contains 3,471 English pairs
across nine domains and 69,420 parallel pairs across 20 languages. All three
protocols agree on the ranking, led by Gemini 2.5 Flash. The 13.5 pp gap
between Gemma4 31B and Sarvam 30B, widening to
17.4--28.3 pp across Indic languages, challenges the assumption that
Indic-native pre-training is enough to beat a capable general-purpose model. However, a fair comparison requires closely matched model size, architecture, and training setup. Law and Sports remain hard everywhere. Two methodological findings follow: the
20--25\% removal rate shows that human oversight stays essential, and the
17--25 pp shortfall of string matching against the judge, largest where answers
are Indic words with more than one accepted spelling, shows that lexical
metrics under-report the very knowledge such benchmarks measure.

\section*{Limitations}

The multilingual evaluation covers 9 of the 19 Indic languages, 2 of the
6 models, and the judge protocol only; conclusions about the rest are limited
to the English subset. Judge verdicts come from a single model (Gemma 3 12B).
The lexical criteria reproduce the ranking, but cannot confirm absolute judge
accuracy, since both are stricter criteria rather than independent ground
truth; a human agreement study is still needed. Gemini 2.5 Flash is both the
generation and evaluation model, introducing a familiarity bias we
cannot fully control for. GPT-5.4 Mini is both the translation model and an
evaluated model; its English-only scores are unaffected, but this matters once
all six are evaluated. The manual review is reported as an aggregate removal
rate, with no inter-annotator agreement statistic. Finally, the 19 non-English
subsets are machine translations of one English pool, so translation errors,
especially in low-resource languages, may affect scores independently of model
knowledge.

\section*{Acknowledgements}

This work was carried out under the L3Cube Labs, Pune mentorship programme. We
thank our mentor for the continuous guidance and support that shaped this work. This work is a part of the L3Cube-IndicNLP project\footnote{\href{https://github.com/l3cube-pune/indic-nlp}{L3Cube-IndicNLP}} \cite{joshi2022l3cube_mahanlp}.

\bibliography{main}

\appendix

\section{Evaluation Pipeline}

\begin{figure}[H]
  \centering
  \includegraphics[width=\columnwidth]{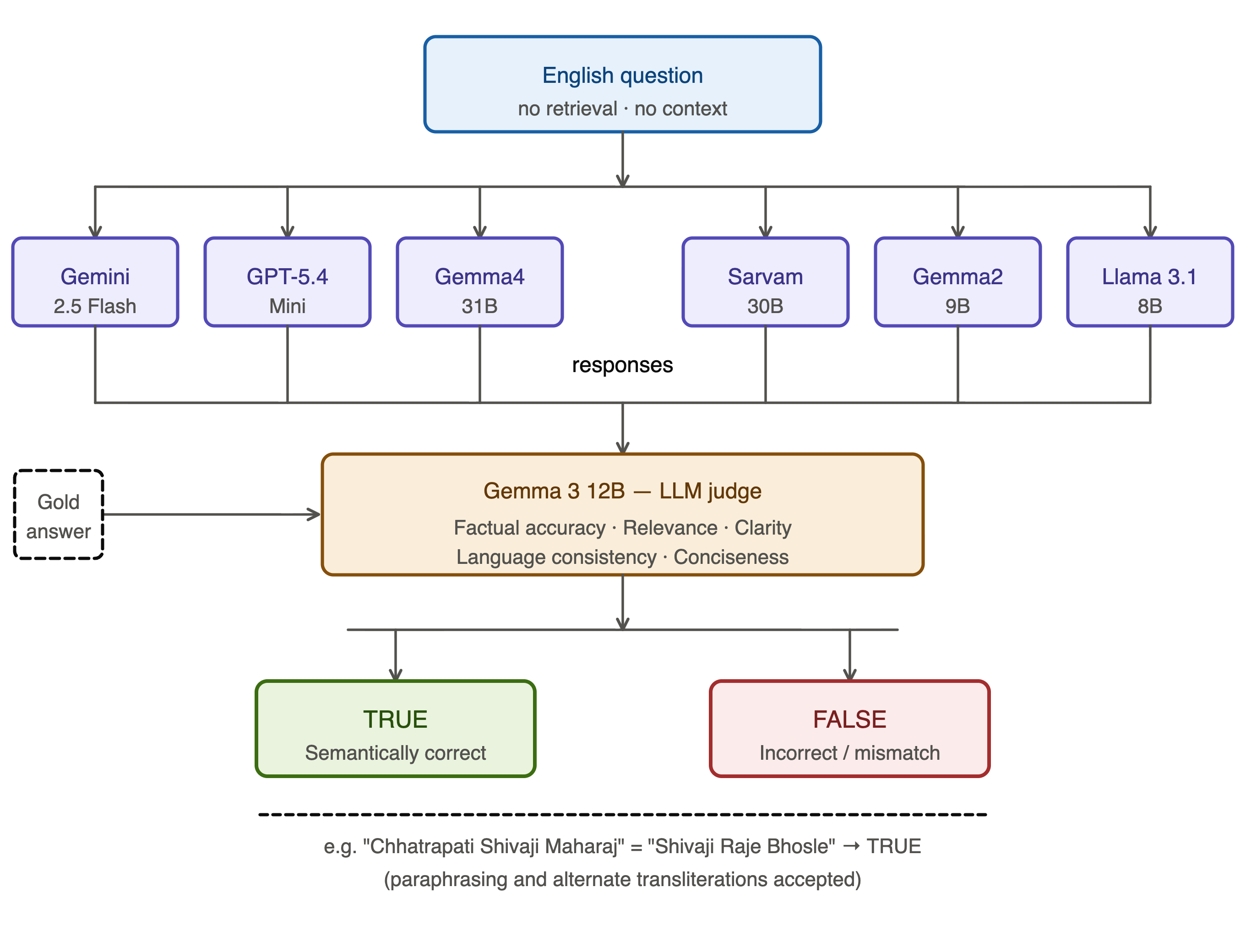}
  \caption{LLM-as-a-judge evaluation pipeline. Each of the six models responds
    to the same question; Gemma 3 12B evaluates each response against the
    gold-standard answer and returns a TRUE/FALSE verdict.}
  \label{fig:eval}
\end{figure}

\section{Domain-wise Lexical Results}

Table~\ref{tab:lexical_domains} gives the full domain-wise scores for both
deterministic criteria defined in Section~\ref{sec:lexical}, on the same
3,471-question English subset used for Table~\ref{tab:heatmap}.

\begin{table*}[t]
  \centering
  \footnotesize
  \renewcommand{\arraystretch}{1.05}
  \setlength{\tabcolsep}{6pt}
  \begin{tabular}{lcccccccccr}
    \toprule
    \textbf{Model} & \textbf{Art} & \textbf{Comm.} & \textbf{Cult.} &
    \textbf{Geo.} & \textbf{Hist.} & \textbf{Law} & \textbf{Pol.Sci.} &
    \textbf{Sci.} & \textbf{Sports} & \textbf{Combined} \\
    \midrule
    \multicolumn{11}{l}{\textit{(a) Exact substring}} \\
    Gemini 2.5 Flash & 49.8 & 68.0 & 43.8 & 53.7 & 66.5 & 85.6 & 68.9 & 55.7 & 64.5 & \textbf{61.97} \\
    GPT-5.4 Mini     & 37.3 & 49.7 & 35.2 & 32.3 & 33.1 & 41.3 & 48.0 & 36.3 & 33.7 & \textbf{37.77} \\
    Gemma4 31B       & 39.9 & 45.1 & 36.3 & 30.7 & 32.8 & 43.5 & 46.4 & 34.8 & 30.4 & \textbf{37.34} \\
    Sarvam 30B       & 25.3 & 39.9 & 23.4 & 22.0 & 24.5 & 22.9 & 40.9 & 34.8 & 26.4 & \textbf{27.72} \\
    Gemma2 9B        & 19.7 & 30.1 & 19.1 & 20.6 & 15.1 & 19.2 & 34.2 & 24.3 & 21.6 & \textbf{21.92} \\
    Llama 3.1 8B     & 16.3 & 29.4 & 15.2 & 18.0 & 12.0 & 16.0 & 28.9 & 17.9 & 18.7 & \textbf{18.24} \\
    \midrule
    \multicolumn{11}{l}{\textit{(b) Word overlap}} \\
    Gemini 2.5 Flash & 48.5 & 67.3 & 43.3 & 54.5 & 65.9 & 85.4 & 71.3 & 57.5 & 64.1 & \textbf{62.23} \\
    GPT-5.4 Mini     & 36.0 & 48.4 & 34.6 & 32.1 & 32.4 & 41.3 & 48.7 & 37.5 & 33.0 & \textbf{37.54} \\
    Gemma4 31B       & 38.6 & 45.1 & 36.0 & 30.9 & 32.0 & 43.3 & 49.3 & 35.1 & 30.0 & \textbf{37.45} \\
    Sarvam 30B       & 24.0 & 39.9 & 23.0 & 21.2 & 24.3 & 22.5 & 44.4 & 35.4 & 26.0 & \textbf{27.86} \\
    Gemma2 9B        & 18.4 & 29.4 & 18.5 & 20.0 & 14.5 & 18.2 & 34.4 & 24.3 & 21.6 & \textbf{21.43} \\
    Llama 3.1 8B     & 15.9 & 28.8 & 14.6 & 17.4 & 11.4 & 15.0 & 30.0 & 17.9 & 18.3 & \textbf{17.89} \\
    \bottomrule
  \end{tabular}
  \caption{Domain-wise accuracy (\%) under the two deterministic criteria,
    English subset. Combined is the micro-average over all 3,471 questions.
    Rows are ordered by combined exact-substring score.}
  \label{tab:lexical_domains}
\end{table*}

\section{Agent and Evaluation Prompts}
\label{app:prompts}

Only the structure of each prompt is shown; rule bodies, worked examples, and
JSON schemas are abridged (marked \texttt{[\ldots]}). Template variables in
double braces (e.g., \texttt{\{\{ context \}\}}) are substituted at runtime.
Prompts are reproduced verbatim from the implementation, and their spelling
follows the original.

\subsection{Agent 1: QA Generation Prompt}
\label{app:agent1}

\begin{lstlisting}[style=promptbox]
You are an expert quizmaster AI specialized in generating high-quality, India-specific trivia questions for training datasets. Your sole purpose is to create context-rich, unambiguous QA pairs that test deep knowledge about India.

CRITICAL REQUIREMENTS - Follow these rules STRICTLY:
1.  INDIA-SPECIFIC FOCUS
2.  CONTEXT-RICH QUESTIONS
3.  ABSOLUTE ANSWERS
4.  COMPLETE NAMES
5.  HIGH DIFFICULTY
6.  SHORT ANSWERS (1-5 words, verbatim from text)
7.  NO TEXT REFERENCES
8.  PERFECT GRAMMAR
9.  GOLDEN DATASET QUALITY
10. JSON OUTPUT ONLY
11. SPECIFIC DETAILS
12. UNIQUE KNOWLEDGE
13. CLEAR QUESTION STRUCTURE (What/Which/Who/When/Where/How)
14. VERBAL CONSISTENCY
15. PRECISE TERMINOLOGY
16. QUALITY OVER QUANTITY (return error JSON over low-quality pairs)
17. AVOID TECHNICAL TRIVIA
18. AVOID VAGUE DIRECTIONAL QUESTIONS
19. AVOID INCOMPLETE ANSWERS
20. AVOID AWKWARD PHRASING
21. FOCUS ON SIGNIFICANT KNOWLEDGE

EXAMPLES OF GOOD QA PAIRS:
EXCELLENT EXAMPLES: [...]
BAD EXAMPLES (avoid these): [...]

QUALITY CHECKLIST - Before generating, ensure each question: [...]

JSON FORMATTING RULES:
If you can generate questions:
[{"question": "...", "answer": "..."}]
If you CANNOT generate any questions that meet all criteria:
{"error": "No India-specific difficult questions with short answers could be generated."}

GENERATION INSTRUCTIONS: [...]

TEXT:
{{ context }}
\end{lstlisting}

\subsection{Agent 2: Quality Validation Prompt}
\label{app:agent2}

\begin{lstlisting}[style=promptbox]
You are an ULTRA-STRICT validator for India-specific QA triplets that enforces GOLDEN TRUTH standards. Your role is to ensure every QA pair meets the highest quality benchmarks for testing LLM knowledge about India.

Input variables:
- source_text: JSON array of {"chunk_id", "chunk_text"}
- qa_triplets: JSON array of {"chunk_id", "question", "answer"}

TASK
1. Find the matching chunk_text using chunk_id.
2. Apply ULTRA-STRICT validation against golden truth standards.
3. Assign quality scores (correctness, relevance, difficulty).
4. Check duplicates across all triplets.
5. Verify GOLDEN TRUTH via Google Search with multiple sources.

ULTRA-STRICT VALIDATION RULES (ZERO TOLERANCE FOR VIOLATIONS)
1.  INDIA-SPECIFIC FOCUS (CRITICAL)
2.  CONTEXT-RICH & SELF-CONTAINED (CRITICAL)
3.  ABSOLUTE ANSWER REQUIREMENTS (CRITICAL; answer verbatim in source, 1-5 words)
4.  PERFECT GRAMMAR & STRUCTURE (CRITICAL)
5.  HIGH DIFFICULTY & SPECIALIZED KNOWLEDGE (CRITICAL)
6.  NO TEXT REFERENCES (CRITICAL)
7.  GOLDEN TRUTH VERIFICATION (CRITICAL; sets "correctness_flag")
8.  QUALITY STANDARDS (CRITICAL)
9.  DUPLICATE DETECTION (BATCH-LEVEL; same answer AND >=80% question overlap)
10. COMPREHENSIVE SCORING SYSTEM (Correctness / Relevance / Difficulty / Overall Quality, each 0-10)
11. DETAILED SCORING GUIDELINES (0-10 SCALE) [...]
    - Overall Quality = Correctness (40%) + Relevance (30%) + Difficulty (30%)
12. EXAMPLES OF VALIDATION [...]
13. ULTRA-STRICT VALIDITY
    - "valid": true ONLY if ALL criteria pass AND "correctness_flag": true
    - Minimum overall quality score of 7 required for validity

Output schema (exact JSON):
{
  "results": [{"chunk_id", "question", "answer", "valid", "correctness_flag", "errors", "scores": {...}, "detailed_assessment": {...}, "duplicate_indices"}],
  "batch_summary": {"total_triplets", "valid_triplets", "invalid_triplets", "duplicate_groups", "average_scores": {...}}
}

Input:
source_text: {{ source_text | tojson(indent=2) }}
qa_triplets: {{ qa_triplets | tojson(indent=2) }}
\end{lstlisting}

\subsection{Model Evaluation Prompt}
\label{app:qa_prompt}

\begin{lstlisting}[style=promptbox]
You are a question-answering assistant. Your task is to answer questions accurately and concisely.

## Input Format
[{"QID": "string", "Question": "string"}, ...]

## Output Format
[{"QID": "string", "Answer": "string", "Reasoning": "string (optional)"}, ...]

## Instructions
1. Answer Accuracy: based on your knowledge.
2. Answer Length: 1 to 5 words maximum.
3. Unknown Answers: set "Answer" to null. Do not make up answers.
4. Reasoning: optional.
5. Output Format: return ONLY the JSON array.
6. Output Language: answer ONLY in the language of the QUESTION.

## Example [...]

Now, answer the following questions:
{{ questions_json }}
\end{lstlisting}

\subsection{LLM-as-a-Judge Prompt}
\label{app:judge_prompt}

\begin{lstlisting}[style=promptbox]
You are an evaluation judge. Your task is to determine if the given answer matches the ground truth answer for the given question.

## Input
Question: {{ question }}
Ground Truth Answer: {{ ground_truth_answer }}
Given Answer: {{ llm_answer }}

## Instructions
1. Compare the "Given Answer" against the "Ground Truth Answer" in the context of the question.
2. Correct if semantically equivalent (same meaning / same entity or fact).
3. Minor phrasing, formatting, or extra detail is acceptable if the core answer is correct.
4. Clearly wrong, meaning-changing, or unrelated answers are incorrect.
5. Answers in a different language than the question and ground truth are incorrect.

## Output Format
Return ONLY a JSON object (no additional text, no markdown fences):
{"is_correct": true, "reasoning": "brief explanation"}
\end{lstlisting}

\end{document}